\documentclass{article} % For LaTeX2e
\usepackage{iclr2016_conference,times}
\usepackage{hyperref}
\usepackage{url}
\usepackage{graphicx}
\usepackage{pdfpages}
\usepackage{amsmath}

\title{How much data is needed to train a medical image deep learning system to achieve necessary high accuracy?}

\author{Junghwan Cho, Kyewook Lee, Ellie Shin, Garry Choy, and $^\dagger$Synho Do
\thanks{Address: 25 New Chardon Street Suite 400B, Boston, MA, 02114, USA, Email: sdo@mgh.harvard.edu, This project is approved by Institutional Review Board of Massachusetts General Hospital. } \\
Department of Radiology\\
Massachusetts General Hospital and Harvard Medical School\\
Boston, MA, USA \\
}

\begin{document}

\maketitle

\begin{abstract}
The use of Convolutional Neural Networks (CNN) in natural image classification systems has produced very impressive results. Combined with the inherent nature of medical images that make them ideal for deep-learning, further application of such systems to medical image classification holds much promise. However, the usefulness and potential impact of such a system can be completely negated if it does not reach a target accuracy. In this paper, we present a study on determining the optimum size of the training data set necessary to achieve high classification accuracy with low variance in medical image classification systems. The CNN was applied to classify axial Computed Tomography (CT) images into six anatomical classes. We trained the CNN using six different sizes of training data set ($5$, $10$, $20$, $50$, $100$, and $200$) and then tested the resulting system with a total of 6000 CT images. All images were acquired from the Massachusetts General Hospital (MGH) Picture Archiving and Communication System (PACS). Using this data, we employ the learning curve approach to predict classification accuracy at a given training sample size. Our research will present a general methodology for determining the training data set size necessary to achieve a certain target classification accuracy that can be easily applied to other problems within such systems.
\end{abstract}

\section{Introduction}

The rapidly increasing amount of medical images and modalities that must be considered presents a growing risk for human error and delayed diagnosis. The current answer to these concerns remains the computer aided detection (CADe) and diagnosis (CADx) systems. However, despite recent development and improvement of these systems, they remain limited in their potential by their use of hand-crafted features (\cite{shiraishi2011computer}). This limitation suggests the potential of a deep-learning approach, which would rather allow the system to extract these features itself (\cite{jones2014learning}). However, while recent years have shown significant advances in image classification problems using Convolutional Neural Networks (CNN), its potential in medical applications has only been minimally explored (\cite{ypsilantis2015predicting}, \cite{hua2015computer}).

Large medical image data sets within a hospital Picture Archiving and Communication System (PACS) combined with advanced high performance parallel computing promises the capacity to accelerate a machine learning technique to more accurately detect clinical imaging findings and diagnose specific diseases (\cite{roth2015improving}, \cite{roth2015anatomy}). This potential is heightened by the qualifications and characteristics of medical images that make them so ideal for deep learning.

However, while medical images present an ideal format and structure for deep learning, it is difficult to secure a large quantity of such medical images because of patient privacy and security policies, such as the Health Insurance Portability and Accountability Act (HIPAA) privacy rule. Regardless of such limitations, such a platform presents an uncompromisable necessity to be of the highest accuracy, consistent in its performance, and fast, as its functionality is directly linked to human life.

Together, these concerns present the crucial question of \textit{how much data is needed to train a medical image deep learning system to achieve necessary high accuracy}. This key question was not explored systematically in the recent medical image deep learning publications (\cite{anavi2015comparative}, \cite{havaei2015brain},\cite{bar2015deep}, \cite{roth2015anatomy}, \cite{yan2015bodypart}, \cite{roth2015detection}, \cite{shin2015interleaved}, \cite{wang2015neutrophils}, \cite{xu2015deep}, \cite{srivastava2015using}, \cite{srivastava2015using}, \cite{zhang2015deep}, \cite{margeta2015fine}.

Several approaches to this question have been introduced and explored in different applications (\cite{figueroa2012predicting}, \cite{beleites2013sample}, \cite{dobbin2007sample}, \cite{dobbin2008large}). Upon thorough evaluation and consideration of these approaches, we chose the learning curve method due to its shown promise and robustness within other applications (\cite{figueroa2012predicting}).

In this paper, we present our initial results from applying deep learning to medical image classification and our approach to determining the ideal training data size to achieve high accuracy. While the learning curve will vary for different problems within this application, our work presents a general methodology that can be easily applied to such different problems to generate learning curves and determine respective necessary target data set sizes.

\begin{figure}[tbp]
\begin{center}
%\framebox[4.0in]{$\;$}
%\fbox{\rule[-.5cm]{0cm}{4cm} \rule[-.5cm]{4cm}{0cm}}
%\includegraphics[width=4in]{Fig1_whole_body_scan.pdf}
\includegraphics{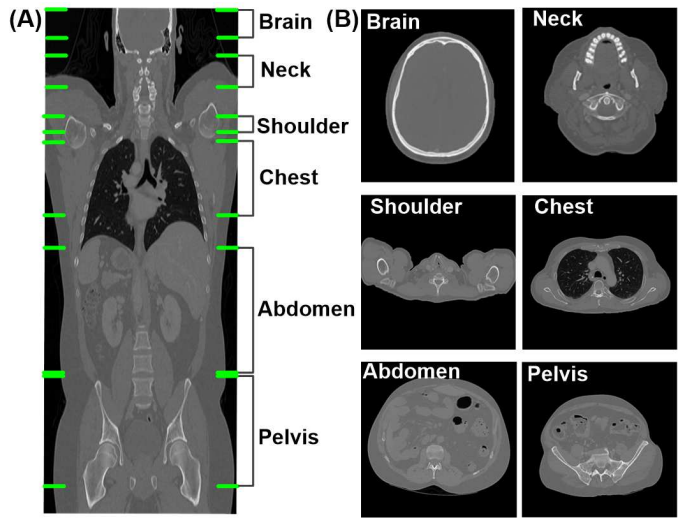}

\end{center}
\caption{: (A) Images of whole-body CT scan and region of each body part (B) examples of representative image of six classes of anatomy.}
\label{Fig1-whole-body}
\end{figure}

\subsection{Medical Image Uniqueness}

Compared to natural images, medical images have unique characteristics well fitted to machine learning.
The potential of medical image deep learning platforms is heightened by the unique inherent characteristics of medical images that make them ideal for machine learning.

First, all the images are standardized. Medical images are acquired from many different modalities (i.e., X-ray radiography, Mammography, Ultrasound, Computed Tomography (CT), Magnetic Resonance Image (MRI), Positron Emission Tomography (PET), Single Photon Emission Computed Tomography (SPECT) etc). However, all the resulting images must be saved and archived in a single format, the Digital Imaging and Communications in Medicine (DICOM) format. DICOM file meta data include the field of view of image, matrix size, pixel pitch, slice thickness, imaging modality, bit resolution, scan protocol, patient information, etc. DICOM files are easy to handle, store, print, and transmit because they include a file format definition and a network communications protocol. Resultingly, the DICOM format is used by all medical imaging systems to generate images and by Picture Archiving and Communication System (PACS) to achieve and retrieve images.

Additionally, not only are all medical images standardized by format, they are also regulated for high quality. Since the quality of medical images directly affects patient safety and quality of healthcare, all hospitals have a quality control division or committee to constantly ensure that the image quality meets the required compliance.

Lastly, all images come with manually annotated radiologist's reports which are structured to report their findings, impressions, and diagnoses, saved in the Electronic Health Record (EHR). These reports present an extremely useful resource for supervised medical image machine learning. However, it is difficult to obtain access to these images as medical images are protected by the Health Insurance Portability and Accountability Act (HIPAA) security rule, which protects a patient's medical records and other personal health information.

As a result, medical data sets are stored within the secure storage system in a hospital, and strict guidelines are put in place even after an anonymization process. Despite the availability of government research center repositories with medical images available for the use of healthcare research (i.e., National Biomedical Imaging Archive (NBIA), Digital Database for Screening Mammography (DDSM), Digital Retinal Images for Vessel Extraction (DRIVE) etc), it remains extremely difficult to find a database with millions of cases.

\section{Material and Method}

\subsection{Data Acquisition}
We first compiled a database of the CT images of patients from the clinical PACS of MGH with an Institutional Review Board (IRB) approval. We developed a preprocessing software to annotate and categorize these images into which body part is represented: brain, neck, shoulder, chest, abdomen, and pelvis.

As can be seen in Figure \ref{Fig1-whole-body}(A), we only utilized the scans of regions that could be clearly defined as one of the aforementioned body parts. The gaps account for transition regions, which were not used for a training algorithm due to their lack of clear regional definition.

In the axial CT image, each scan has different noise level because of different radiation dosage and different image reconstruction filters and different CT vendors.  Also, each image voxel has different pitch because of image reconstruction fields are different. The slice thickness of each image are thicker than axial voxel pitch so voxel is anisotropic usually axial pixel pitch is (i.e., $dx = dy = 0.3 \sim 0.6mm$) and slice thickness varies by image reconstruction parameters. (i.e., $dz = 3 \sim 5mm$)

\subsection{Convolutional Neural Network}

The body part classification problem is an important step in initial classification process for eventual disease classification and abnormal area detection problems. In addressing this problem, several issues must be considered. First, anatomical structures and the details of body regions vary dramatically between patients (i.e., complex brain areas, small bone structures in the neck, small lung nodules and thin pleura in the chest, closely located organs in the abdomen region). Also, major bone structures in the shoulder and pelvis lead to artifacts on medical images. As a result, it is possible that a huge training data set will be necessary for our algorithm to achieve the desired accuracy.

Accordingly, the processing of such big data necessitates a robust algorithm. Over the years, many deep convolutional neural network algorithms have been proposed for natural image classification, such as GoogLeNet (\cite{szegedy2014going}), AlexNet (\cite{krizhevsky2012imagenet}), ClarifaiNet (\cite{zeiler2014visualizing}), VGGNet (\cite{simonyan2014very}) etc. After extensive review of these algorithm, we chose to implement and use GoogLeNet for our specific application because it is computationally efficient and provides high quality classification based on the Hebbian principle and multi-scale processing as compared to other convolutional neural networks (CNN).

The typical CNN framework consists of several convolutional and sub-sampling layers, followed by a fully connected traditional multiple layer perceptron. The GoogLeNet uses 22 convolutional layers including 9 Inception modules and 4 different size of basis or kernel filters (i.e., $7\times7, 5\times5, 3\times3$, and $1\times1$). CNN algorithms can learn the basis vectors of images and extract useful higher-level features through a hierarchical process. Accordingly, feature maps were extracted by convolution from the learned basis vectors of input medical CT images since each body part has different anatomical structures (\cite{roth2015anatomy}). Currently, we are working on improving upon this baseline algorithm to further customize it for medical images.

We then ran the GoogLeNet network using the NVIDIA Deep Learning GPU Training System (DIGITS) DevBox, a deep learning optimized machine with four TITAN GPUs with 7 TFlops of single precision, 336.5 GB/s of memory bandwidth, and 12 GB of memory per board, to train our model using each experimental data set. The convolution filters of the GoogLeNet network were trained using stochastic gradient descent (SGD) algorithms until it ran to 200 training epochs. Validation data sets were presented upon every epoch during the training process. The initial base learning rate was 0.01 and decreased by three steps according to the convergence to object function.

\subsection{Learning Curve}

The classification accuracy of deep-learning classifiers is largely dependent on the size of high quality initial training data sets. As a result, the investigation of how large a training data set is needed to achieve the target accuracy for anatomy image classification is imperative.

The learning curve approach of modeling classification performance as a function of the training sample size can predict the sample size needed to train a certain image classification system. Generally, this curve model is represented as an inverse power law function (\cite{figueroa2012predicting}) The classification accuracy ($\textbf{y}$) is expressed as a function of the training set size ($\textbf{x}$) where given unknown parameter ($\textbf{b}=b_1, b_2$). The learning curve was modeled by the following equation.

\begin{eqnarray}
  \label{eq:learning-curve}
  \textbf{y}=f(\textbf{x}; \textbf{b})=100 + b_1 \cdot \textbf{x}^{b_2}
\end{eqnarray}

where $\textbf{x}=[x_1, x_2, \cdots, x_6]^T$, $\textbf{y}=[y_1, y_2, \cdots, y_6]^T$, and $\textbf{b}=[b_1, b_2]^T$. $b_1$ and $b_2$ represent the learning rate and decay rate respectively (\cite{figueroa2012predicting}). The model fit assumes that the classification accuracy ($\textbf{y}$) grows asymptotically to 100\%, or maximum achievable performance.

Using the observed classification accuracy at six different sizes of training sets ($5$, $10$, $20$, $50$, $100$, and $200$), unknown parameters ($\textbf{b}=[b_1, b_2]^T$) were estimated using weighted nonlinear regression (\cite{jang1997neuro}). Given a set of $m=6$ training data pairs ($\textbf{x}, \textbf{t}) = \{(x_p,t_p): p=1,\cdots ,m\}$, we find the optimal ($\textbf{b}=[b_1, b_2]^T$) that minimizes the sum of squared errors:

%\begin{eqnarray}
\begin{align}
  \label{eq:learning-curve}
  E(\textbf{b}) &=\sum_{p=1}^m w_p \cdot (t_p-y_p)^2\\
  &=\sum_{p=1}^m w_p \cdot (t_p - f(x_p, \textbf{b}))^2 \\
  &=\sum_{p=1}^m w_p \cdot r_p (\textbf{b})^2 \\
  &=\textbf{R}^T W \textbf{R}
\end{align}
%\end{eqnarray}
where $t_p$ is the desired output when input is $x_p$; $y_p = f(x_p;\textbf{b})$ is the model's output when input is $x_p$; $r_p(\textbf{b})$ is the residual between $t_p$ and $y_p$ and $\textbf{R}$ is the matrix form. The weight terms $w_p$ in the diagonal matrix $W$ can be determined by applications. The weighted nonlinear least-squares estimator is more appropriate than a regular nonlinear regression method to fit the learning curve when measurement errors do not all have the same variance (\cite{dennis1981adaptive}).

Since the observed classification accuracy using the larger sizes of training sets (such as $100$ and $200$) had a lower variance than when using smaller sample sizes (such as $5$, $10$, $20$, and $50$), the learning curve was fitted by higher weighting values at the points of larger data set sizes. We chose $w_p =\{1,1,1,1,100,150\}$ in this study but it will be $w_p=\{1,1,1,1,1,1,1\}$ for unweighted nonlinear least-squares estimator.

\begin{figure}[htbp]
\begin{center}
%\framebox[4.0in]{$\;$}
%\fbox{\rule[-.5cm]{0cm}{4cm} \rule[-.5cm]{4cm}{0cm}}
\includegraphics[width=5.5in]{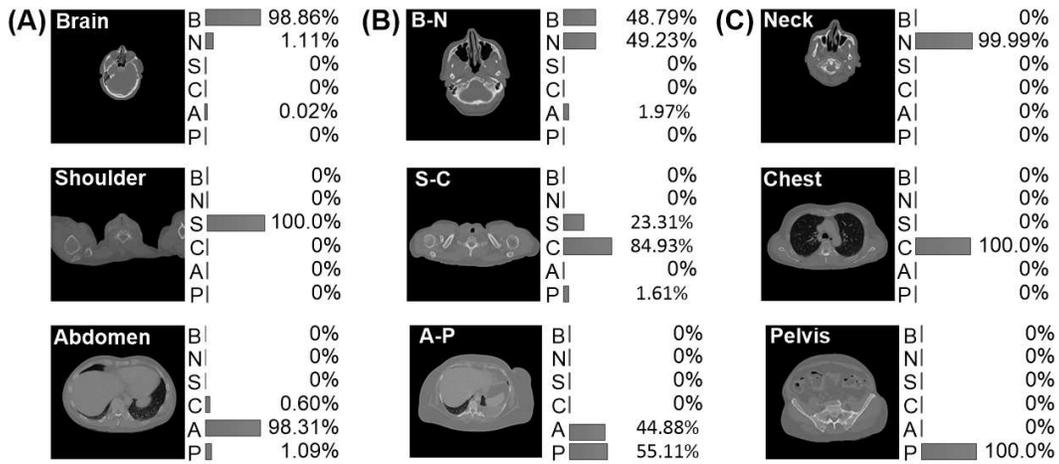}
\end{center}
\caption{Example of new introduced image and resulting accuracy of identification of body part (B: Brain, N: Neck, S: Shoulder, C: Chest, A: Abdomen, P: Pelvis, B-N: image between brain and shoulder, S-C: shoulder and chest, A-P: Abdomen and pelvis). }
\label{Fig2-Test-example}
\end{figure}

\begin{figure}[tbp]
\begin{center}
%\framebox[4.0in]{$\;$}
%\fbox{\rule[-.5cm]{0cm}{4cm} \rule[-.5cm]{4cm}{0cm}}
\includegraphics[width=5.5in]{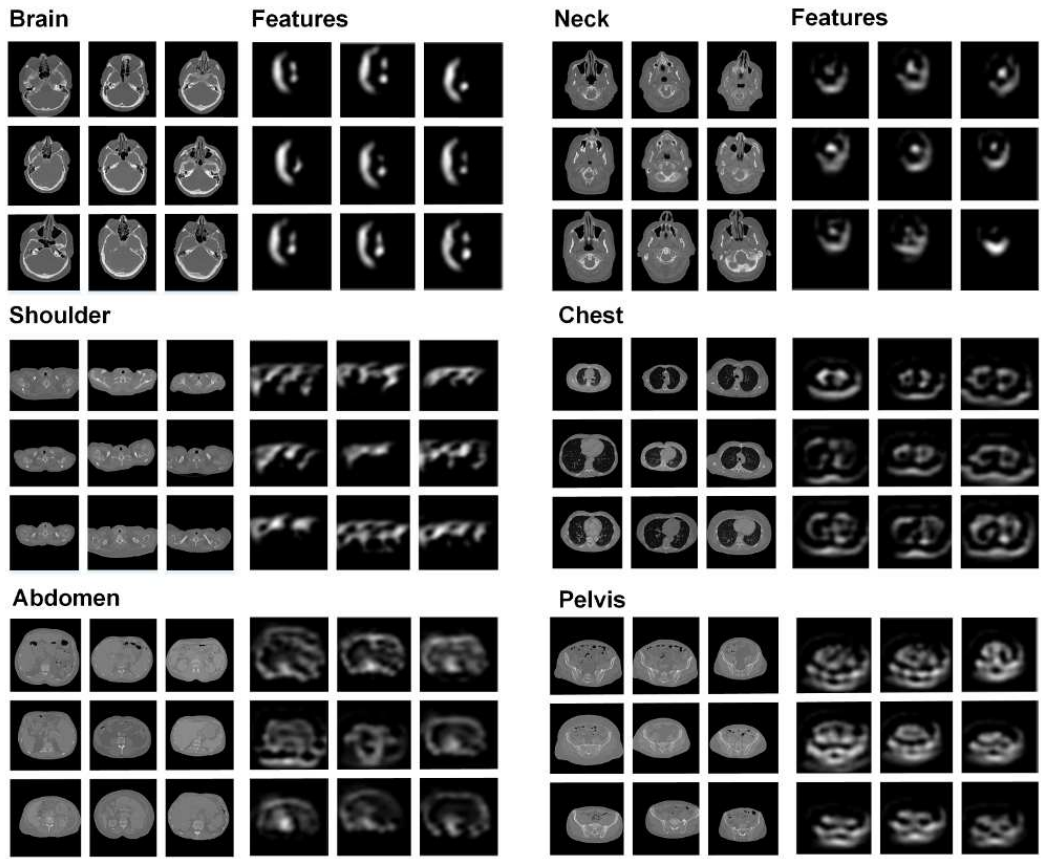}
\end{center}
\caption{Tested CT images and their Inception (3a) modules outputs from GoogLeNet}
\label{Fig3-All-feature}
\end{figure}

\section{Results and Discussions}

Using the Caffe (\cite{jia2014caffe}) and GoogLeNet network implemented on DIGITS DevBox, we conducted six different experiments using varying sizes of training data sets and recorded the resulting classification accuracy. Each experiment was repeated 10 times for each sample size and validated using data sets (25\% randomly selected from each training set). We then tested the trained model by introducing 1000 new images of each body class, a total of 6000 CT images in this study but will be increased continuously.

Figure \ref{Fig2-Test-example} shows representative classification accuracy results for the testing of one repetition of the 200 training size. For all defined body parts, the classification accuracy was near or at 100\%, as shown in Figure \ref{Fig2-Test-example}(A) and (C). In addition, although the system was not trained on any images of transition regions, it was able to infer with considerable accuracy the body parts in transition of such images, as shown in Figure \ref{Fig2-Test-example}(B). This ability of the system to classify images outside of just the six that it was trained on shows immense promise of future development of human anatomy learning algorithms not limited to specific regions but with the ability to classify any image of human anatomy.

In this classification application, a tested image is correctly recognized as the target class if it results in above a 95\% recognition rate. It should also be noted that while the images presented to be tested varied in spatial complexity and variability, the deep learning classifier was still able to correctly identify nearly all of the images based off of feature recognition from its training period.

Figure \ref{Fig3-All-feature} shows tested CT images and their Inception (3a) modules outputs from GoogLeNet. Even though the nine images of the same anatomic region present a wide range of spatial complexity and variability, our trained system was able to extract and identify shared similar features at the level of Inception.

\begin{figure}[htbp]
\begin{center}
%\framebox[4.0in]{$\;$}
%\fbox{\rule[-.5cm]{0cm}{4cm} \rule[-.5cm]{4cm}{0cm}}
\includegraphics[width=2.7in]{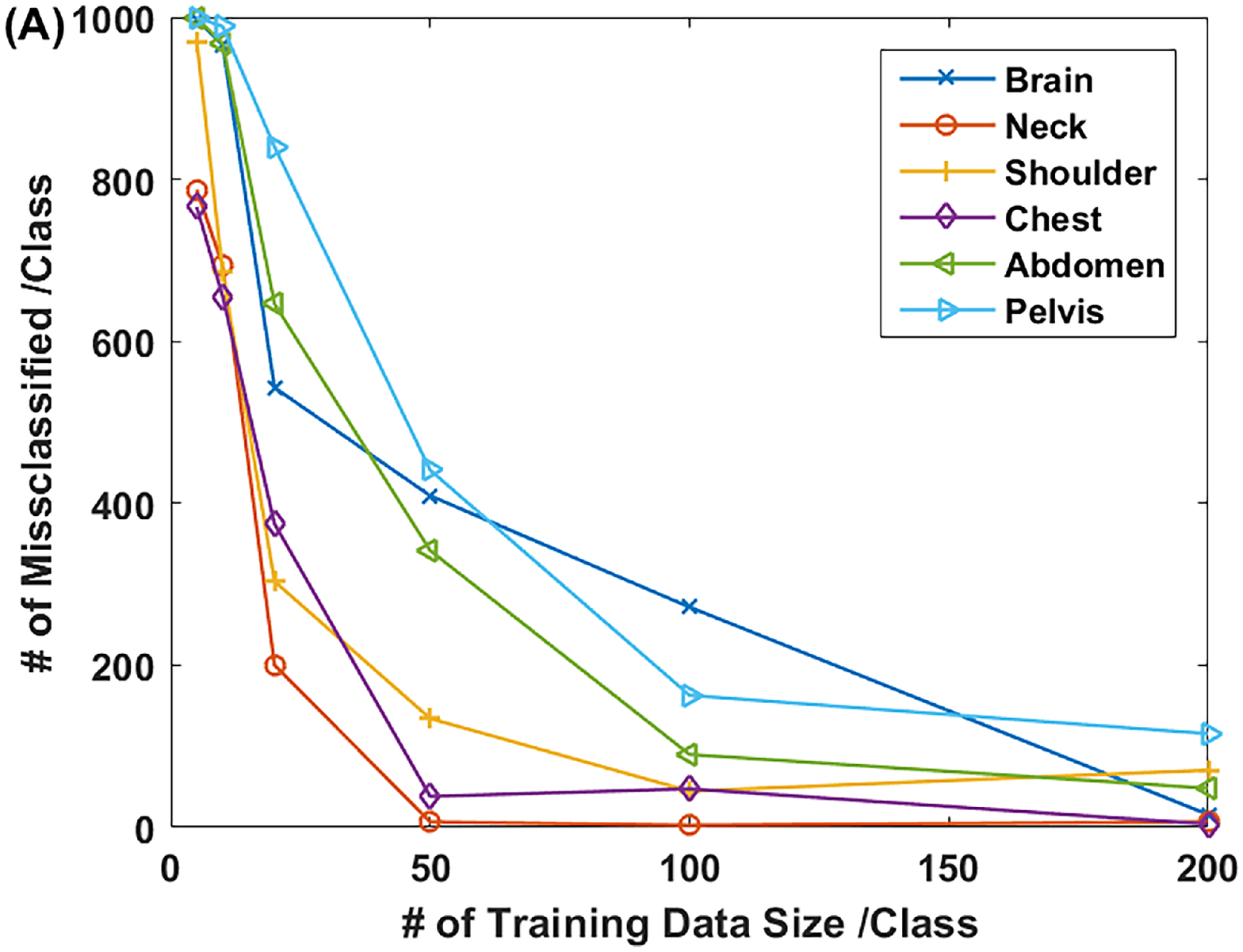}
\includegraphics[width=2.7in]{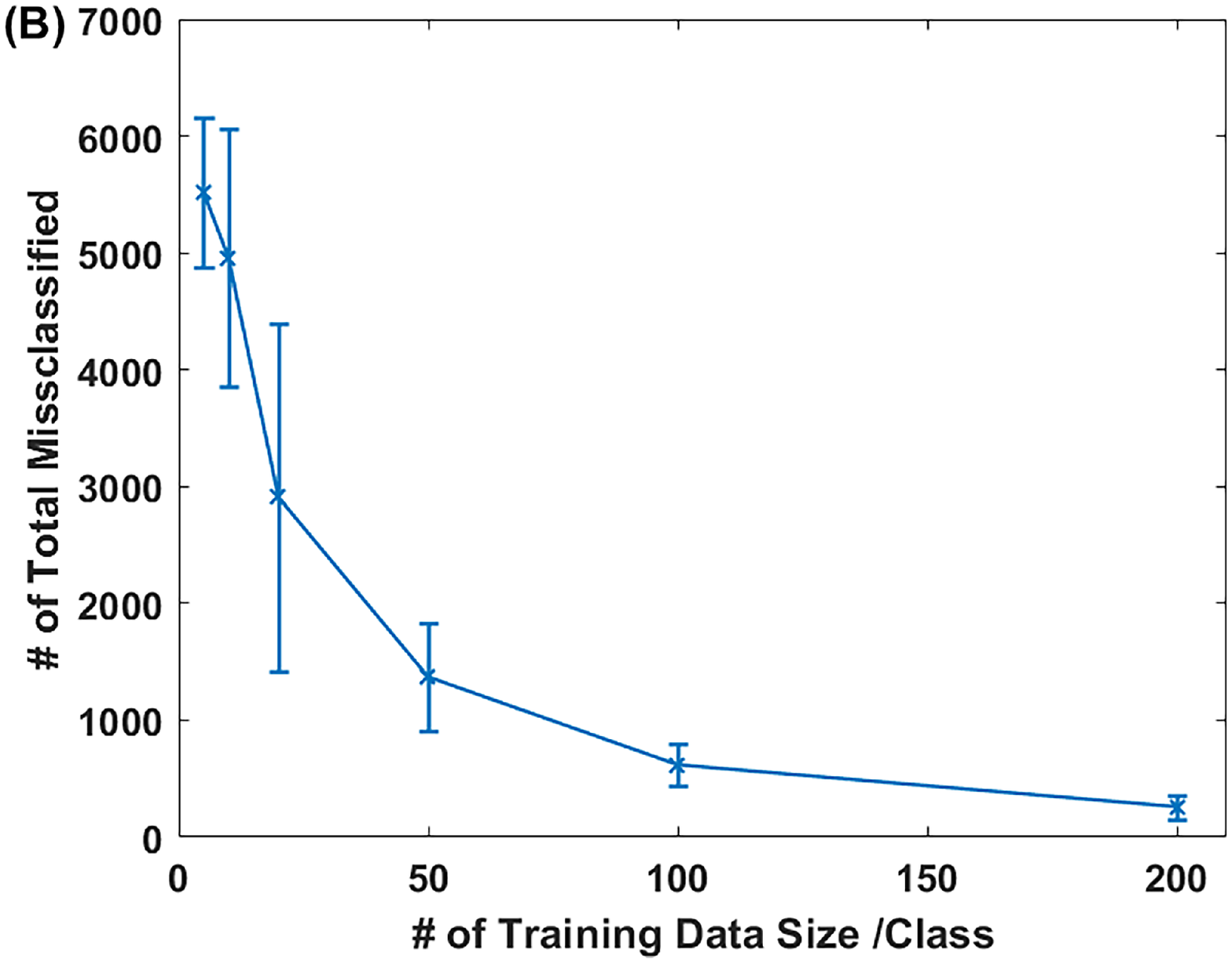}
\end{center}
\caption{(A) The number of misclassified images on each body part class and (B) of total misclassified ones on whole body in increasing number of training data sets.  }
\label{Fig4-misclassified}
\end{figure}

The average results for each body part and the whole body are summarized on Table \ref{Table1-Class-Accuracy}. For all body parts, larger sets of training data led to increased accuracy in classification. This is further evidenced in Figure \ref{Fig4-misclassified}, which graphically shows that the number of misclassified images decreases for each body part with increasingly larger sets of training data. Figure \ref{Fig4-misclassified}(B) also shows that the standard deviation of misclassified images across each experiment repetition decreased with increasing training size except for the 5 and 10 sample sizes. Since too small a sample size led to high misclassification, it had a low standard deviation.

\begin{figure}[htbp]
\begin{center}
%\framebox[4.0in]{$\;$}
%\fbox{\rule[-.5cm]{0cm}{4cm} \rule[-.5cm]{4cm}{0cm}}
\includegraphics[width=2.7in]{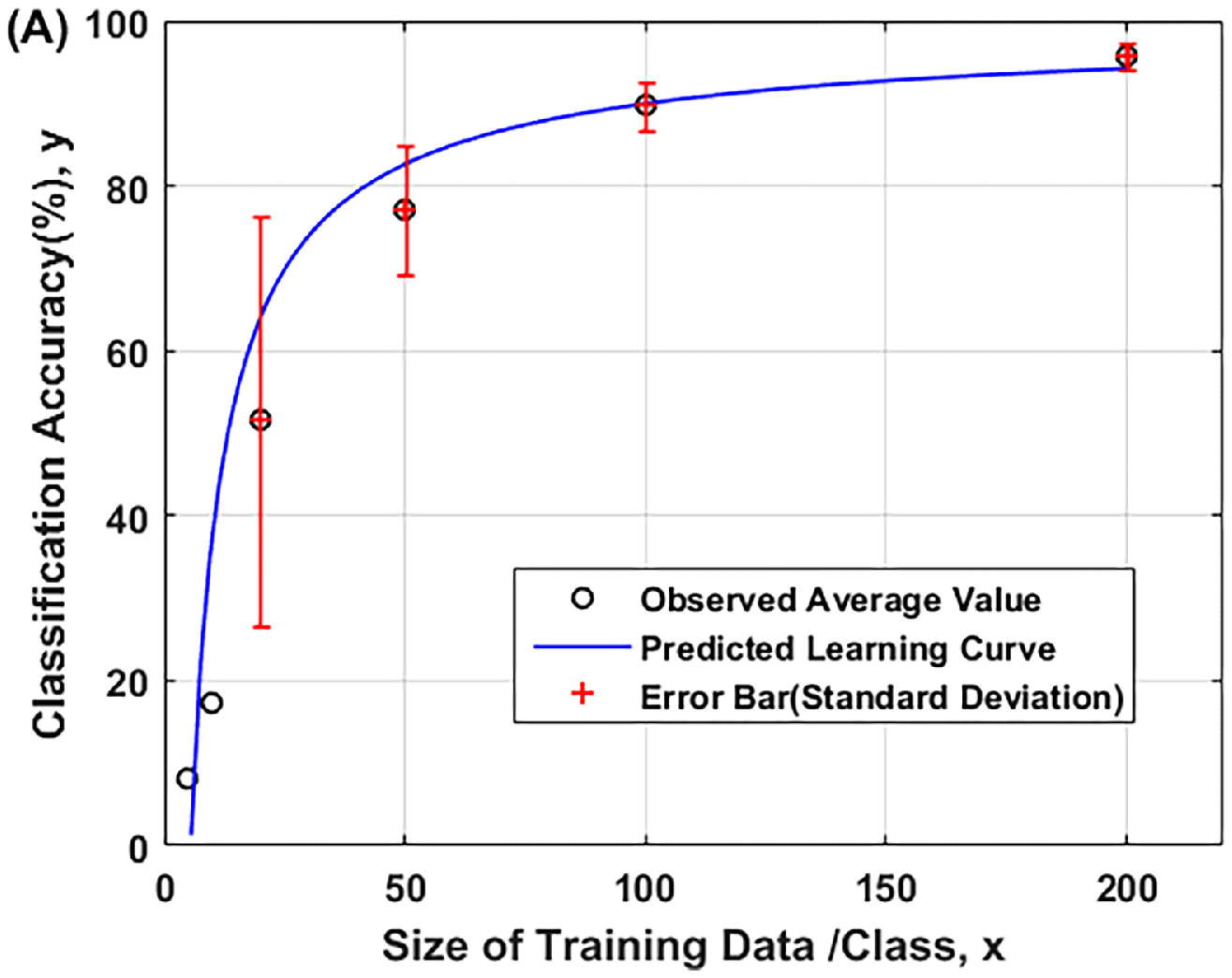}
\includegraphics[width=2.7in]{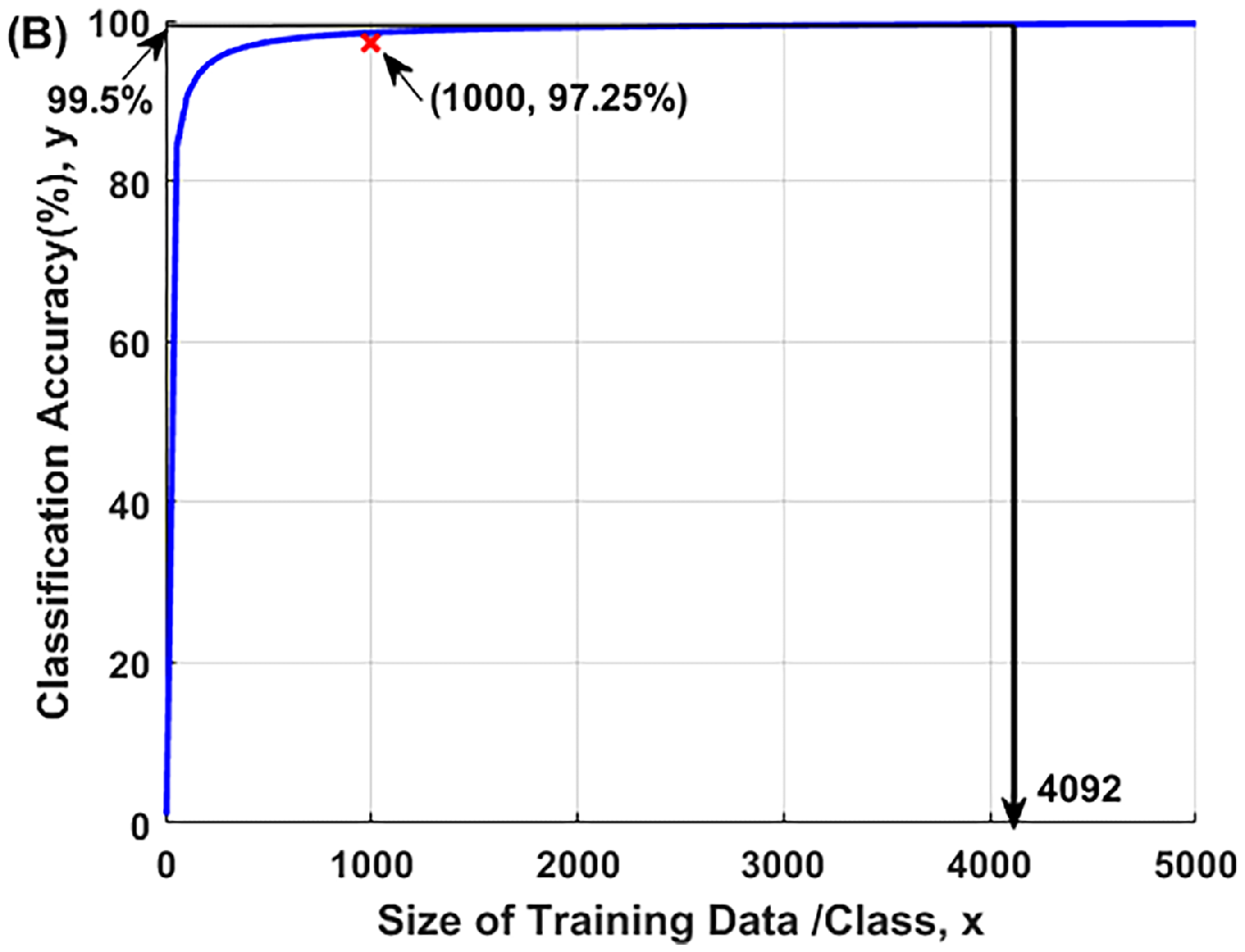}
\end{center}
\caption{(A) The predicted learning curve and (B) tested result at large data set.}
\label{Fig5-learning-curve-test}
\end{figure}

When determining the size of the training data, our primary aim was to achieve high classification accuracy with low standard deviation. The learning curve described in Section 2 was fitted using the observed classification accuracy according to a given training sample size (5, 10, 20, 50, 100, and 200). As shown in Figure \ref{Fig5-learning-curve-test}, the curve much better fit the data points at large sample sizes (100 and 200) by the weighted least square estimator. The classification accuracy increased rapidly from training size 5 to 50, while the accuracy did not increase significantly from training size 100 to 200. After that, the learning curve reached a steady state and did not change much in accuracy regardless of training size.

We also recorded the standard deviation at the six different sizes of training data set. A larger sample size led to a lower standard deviation, exemplified by the predicted classification at the 200 sample size being more accurately and consistently located in the true mean value than at the 20 sample size. The learning curve well predicted a 98\% classification accuracy for the training data size of 1000 per body class, with the observed actual accuracy at 97.25\%. Based on this learning curve, we can predict that our particular deep learning classifier needs a training data set per class of 4092 to reach the desired accuracy, 99.5\%. Accordingly, we will start from a baseline training data set per class size of 5,000 and steadily increase from there in order to achieve a much higher system accuracy.

\section{Conclusion}

While applications of medical image deep learning analysis have only recently begun to be explored, the potentials of such systems are immense due to the unique characteristics of medical images that make them ideal for deep learning. For example, all medical images are regulated for extremely high quality and also manually annotated with a radiologist's report, or a ground truth. These qualities are especially ideal and necessary as such systems require exceptionally high sensitivity and specificity due to their their importance in disease diagnosis and treatment planning. However, patient privacy laws and policies make access to such medical images very difficult, presenting one of the most important questions as we move forward in the field of medical image machine learning - Exactly how large does our training data set have to be to solve a specific classification or detection problem with high accuracy?

We present a learning curve extrapolation method to estimate the required training data size, whose methodology is theoretically well formulated in other fields and similarly provided agreeable results when applied to our system. Further, it can be applied to future problems in medical image deep learning analysis. As the problem changes, the shape of the learning curve will also change.

In order to accurately predict the size and shape of that learning curve, three prerequisites must be met: (1) the training data set must be of high quality, (2) the sampling points must be able to be systematically increased, and (3) each point must be able to be repeated in order to accurately estimate the statistical mean.

Future applications of our work will extend our body part classification to disease classification and organ classification, and also the implementation of abnormal area detection algorithm for each body part. In addition, our approach can be extended to countless other medical imaging modalities.

%Table 1. Classification accuracy results according to increasing size of training data sets
%# of Training Data Size	5	10	20	50	100	200
%Body Part	Brain	0.3	3.39	45.71	59.07	72.82	98.44
%        	Neck	21.3	30.63	79.97	99.34	99.74	99.33
%	        Shoulder	2.98	31.39	69.64	86.57	95.53	92.94
%	        Chest	23.39	34.45	62.53	96.18	95.25	99.61
%	        Abdomen	0.1	3.23	35.4	65.83	91.01	95.18
%	        Pelvis	0	1.15	15.99	55.9	83.71	88.45
%	        Total (averaged)	8.01	17.37	51.54	77.15	89.68	95.67

\begin{table}[t]
\caption{Classification accuracy results according to increasing size of training data sets}
\label{Table1-Class-Accuracy}
\begin{center}
\begin{tabular}{c|c|cccccc}

%\multicolumn{1}{c}{\bf PART}  &\multicolumn{1}{c}{\bf DESCRIPTION}
%\\ \hline \\

\hline
\multicolumn{2}{c}{\bf Training Data Size} 	&\bf 5	&\bf 10	&\bf 20	&\bf 50	&\bf 100	&\bf 200\\
\hline
\hline
Body Part	&Brain	&0.30	&3.39	&45.71	&59.07	&72.82	&98.44\\
        	&Neck	&21.3	&30.63	&79.97	&99.34	&99.74	&99.33\\
	        &Shoulder	&2.98	&31.39	&69.64	&86.57	&95.53	&92.94\\
	        &Chest	&23.39	&34.45	&62.53	&96.18	&95.25	&99.61\\
	        &Abdomen	&0.10	&3.23	&35.40	&65.83	&91.01	&95.18\\
	        &Pelvis	&0	&1.15	&15.99	&55.90	&83.71	&88.45\\
\hline

	        &\bf Average Total	&\bf8.01	&\bf17.37	&\bf51.54	&\bf77.15	&\bf89.68	&\bf95.67\\
\hline
\end{tabular}
\end{center}
\end{table}

\bibliography{iclr2016_conference}

\bibliographystyle{iclr2016_conference}

\end{document}